\documentclass[table]{bmvc2k}
\usepackage{multirow}
\usepackage{bm}
\usepackage{wrapfig}
\usepackage{graphics}
\usepackage{capt-of}
\usepackage{dsfont}

\definecolor{baselinecolor}{gray}{.9}
\newcommand{\cc}{\cellcolor{baselinecolor}}
\newlength\savewidth\newcommand\shline{\noalign{\global\savewidth\arrayrulewidth\global\arrayrulewidth1pt}\hline\noalign{\global\arrayrulewidth\savewidth}}

\title{MixMask: Revisiting Masking Strategy for Siamese ConvNets }

\addauthor{Kirill Vishniakov}{kirill.vishniakov@mbzuai.ac.ae}{1}
\addauthor{Eric Xing}{eric.xing@mbzuai.ac.ae}{1, 2}
\addauthor{Zhiqiang Shen}{zhiqiang.shen@mbzuai.ac.ae}{1}

\addinstitution{
 Mohamed bin Zayed University of \\ Artificial Intelligence
}
\addinstitution{
Carnegie Mellon University
}

\runninghead{Vishniakov et al.}{MixMask}

\begin{document}

\maketitle

\begin{abstract}
The recent progress in self-supervised learning has successfully combined Masked Image Modeling (MIM) with Siamese Networks, harnessing the strengths of both methodologies. Nonetheless, certain challenges persist when integrating conventional erase-based masking within Siamese ConvNets. Two primary concerns are: (1) The continuous data processing nature of ConvNets, which doesn't allow for the exclusion of non-informative masked regions, leading to reduced training efficiency compared to ViT architecture; (2) The misalignment between erase-based masking and the contrastive-based objective, distinguishing it from the MIM technique. To address these challenges, this work introduces a novel filling-based masking approach, termed \textbf{MixMask}. The proposed method replaces erased areas with content from a different image, effectively countering the information depletion seen in traditional masking methods. Additionally, we unveil an adaptive loss function that captures the semantics of the newly patched views, ensuring seamless integration within the architectural framework. We empirically validate the effectiveness of our approach through comprehensive experiments across various datasets and application scenarios. The findings underscore our framework's enhanced performance in areas such as linear probing, semi-supervised and supervised finetuning, object detection and segmentation. Notably, our method surpasses the MSCN, establishing MixMask as a more advantageous masking solution for Siamese ConvNets. 
Our code and models are publicly available at \href{https://github.com/kirill-vish/MixMask}{github.com/kirill-vish/MixMask}.
\end{abstract}

%-------------------------------------------------------------------------
\section{Introduction}
Self-supervised learning is a popular method for deriving representations from data without requiring human-labeled annotations. One common approach within this paradigm is the Siamese Network, characterized by its dual-branch structure. The distance or relationship between these two branches is typically assessed using similarity loss~\cite{grill2020bootstrap,chen2021exploring}, contrastive loss~\cite{oord2018representation,he2020momentum,chen2020simple}, or distillation loss~\cite{caron2021emerging}. A recent innovation, termed Masked Image Modeling (MIM)~\cite{he2022masked,bao2021beit,assran2022masked}, has shown promise in enhancing representation learning. To integrate the strengths of both masked design and Siamese networks, the Masked Siamese Network (MSN) built on ViT was introduced~\cite{assran2022masked}. Parallelly, convolutional networks such as ConvNeXt~\cite{liu2022convnet} and its V2 iteration~\cite{woo2023convnext} have also been making strides, exhibiting strong performance in both labeled and unlabeled learning contexts, underlining the potential of ConvNets. A more recent advance is the Masked Siamese ConvNets (MSCN)~\cite{jing2022masked}, which is an extension of MSN but utilizes a ConvNet-based encoder. This method, however, takes the masking strategy directly from MIM~\cite{he2022masked, bao2021beit}, without customizing it for the unique attributes of Siamese ConvNet. Additionally, both MSN and MSCN lean heavily on the multicrop strategy, as detailed in~\cite{caron2020unsupervised,caron2021emerging}, to counterbalance performance limitations stemming from the erasure function. As an example, while MSN produces ten additional perspectives with a focal mask in each cycle, MSCN creates two supplementary views. This results in an increased computational load during training.

Generally, erase-based masking works particularly well in tandem with the image patchify mechanism seen in ViT~\cite{dosovitskiy2020image}. This mechanism creates image patches that the encoder processes independently. By doing so, masked patches can easily be omitted, subsequently reducing computational expenses~\cite{he2022masked}. In contrast, ConvNets employ a continuous data processing approach. This is not suitable for omitting masked patches because it would maintain the same processing time, even though the masked patches no longer carry significant semantic value. Another challenge arises from numerous contrastive frameworks employing embedding loss~\cite{chen2020simple, he2020momentum}. This loss is not structured to restore the removed regions but rather to distinguish between different inputs. Moreover, multiple studies~\cite{cole2022does, chen2021intriguing} have shown that contrastive loss tends to capture more general, coarse-grained features. Masking operations compromise these overarching features, subsequently slowing down a model's convergence rate. In contrast, vision transformer-based MIM methods~\cite{he2022masked,bao2021beit} aim to restore the masked sections by working directly within the pixel domain using a reconstruction loss.

We have pinpointed two primary limitations associated with Masked Siamese ConvNets: {\bf (1)} The standard erase-based masking operation impinges upon global features crucial for the contrastive objective as referred in \cite{chen2021intriguing, cole2022does}. Plainly put, conventional masking excises significant semantic data from the input, and such loss is irretrievable through subsequent processing. To illustrate, masking 25\% of the image means a corresponding 25\% information loss during training, thereby hindering the training's efficacy. This highlights the pressing need for a refined masking approach to optimize the learning of representations. {\bf (2)} The inherent ``symmetric'' semantic distance loss in Siamese networks does not adjust for the shift in semantic distance across varying views of an identically masked image. Notably, the global semantic essence of an image is altered when random segments are erased. Take, for example, an image of a dog in the center. The semantics of this image diverge markedly from a version where distinct portions of the dog are masked and thus hidden. Such uniform loss configuration has a marginal impact when both branches undergo masking. Nonetheless, issues arise when solely one branch gets masked. Given that asymmetric Siamese networks, as proposed in \cite{shen2022unmix}, are identified to be superior learners, a more adaptable loss system utilizing soft distance becomes indispensable to truly capture the semantic gap between the duo of asymmetric branches in masked Siamese convnets.

\begin{figure*}[ht]
  \centering
  \includegraphics[width=0.98\textwidth]{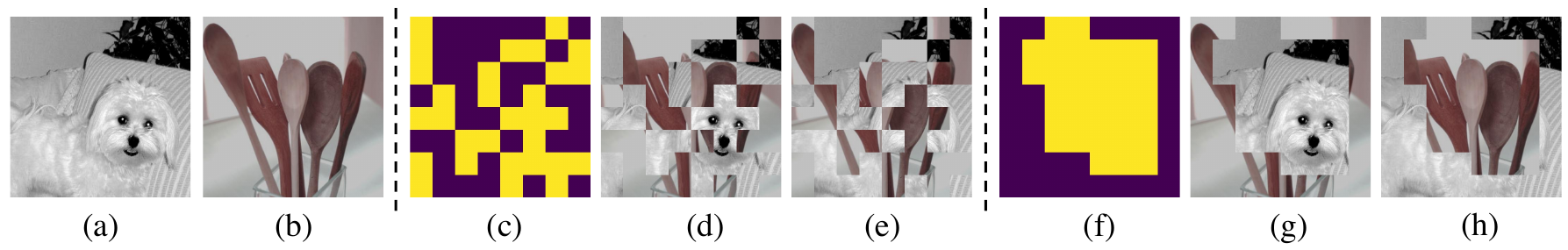}
  \caption{Illustration of the different mask patterns with a mask grid size of 8. (a) and (b) are input images. (c) is the discrete/random mask pattern, and (d) and (e) are mixed images using this mask. (f) is the blocked mask pattern, and (g) and (h) are mixed images with a blocked mask. {Discrete masking breaks (c) -- (e) the completeness of an object which is important for the contrastive loss because it operates on the global object level. On the other hand, blocked masking (f) -- (h) preserves important global features leading to superior performance.}}
  \label{fig:masking_all}
\end{figure*}

To counter the limitations mentioned, we introduce a filling-based masking technique as an alternative to the erasing method, ensuring that there is no information loss. Rather than randomly removing regions of an image, we opt to randomly select another image, harnessing its pixel values to populate the masked segments. Importantly, in contrast to the discrete random masking employed in~\cite{he2022masked}, we discovered that a block-wise approach yields superior results. This method maintains crucial global information, essential for contrastive loss. Paired with our mask-filling strategy, this block-wise masking effectively replaces blank spaces with comprehensive viewpoints from other images within a batch. In recognition of the semantic shift post-filling, we have incorporated the asymmetric loss design as detailed in~\cite{shen2022unmix}. Our filling-based technique, compared to MSCN's erase-based strategy, supplants nondescript erased areas with regions that offer richer semantic insights. Notably, this method does away with the need for multicrops, only demanding one supplementary view per image, marking a significant efficiency leap over MSN and MSCN that utilize ten and two cropping pairs respectively. Such invaluable information derived from our filling-based approach can be harnessed by Siamese networks, amplifying the training representation. Our ablation studies confirm its superior efficacy for the Masked Siamese ConvNets.

Our proposed approach has been exhaustively validated across the CIFAR-100, Tiny-ImageNet, and ImageNet-1K datasets and seamlessly incorporated into various Siamese ConvNets, including MoCo, BYOL, SimCLR, and SimSiam. Across all datasets, our method substantially elevated the performance of multiple baseline models. We also noted consistent improvement in our refined models across semi-supervised and supervised fine-tuning tasks, including object detection and segmentation. Our contributions in this work are as follows:
\vspace{-0.05in}
\begin{itemize}
\addtolength{\itemsep}{-0.05in}
\item {We reveal that {\em Siamese ConvNets} experience reduced convergence when using the conventional erase-based masking scheme. In response, we introduce an effective filling-based masking strategy, which better complements the image-level contrastive objective of self-supervised Siamese ConvNets.}

\item  {We integrate a flexible loss structure with a soft distance to harmonize the combined {\em masking} and {\em Siamese} architecture. This ensures there are no discrepancies between the transformed input and objectives in Masked Siamese ConvNets.}

\item {Extensive experiments are performed on various datasets and Siamese frameworks to solidify the efficacy and broad applicability of our approach. Our results consistently indicate that our method outperforms MSCN in areas such as linear probing, semi-supervised fine-tuning, and downstream tasks like detection and segmentation.}

\end{itemize}

\section{Related Work}

\noindent{\textbf{Masked Siamese Networks.}} 
Recent developments in MIM and siamese self-supervised learning have prompted research efforts to combine these two techniques. One such approach, Masked Siamese Networks (MSN), was proposed in~\cite{assran2022masked}. MSN generates an anchor view and a target view and applies masking operations exclusively to the anchor branch. To assign masked anchor representations to the same cluster as the unmasked target, the method employs prototype cluster assignment. 

By contrast, CNNs are less compatible with masked images as they operate at the pixel level, which can cause image continuity to be disrupted by masking operations. To address this, Masked Siamese ConvNets were introduced in~\cite{jing2022masked}.

\noindent{\textbf{Self-supervised Learning.}}  
Self-supervised learning (SSL) is a technique for representation learning that leverages vast amounts of unlabeled data. Early SSL approaches in the field of computer vision were based on pre-text tasks~\cite{noroozi2016unsupervised, gidaris2018unsupervised, sermanet2018timecontrastive, zhang2016colorful}. A pivotal advancement in SSL was the introduction of a simple contrastive learning framework by~\cite{chopra2005learning,chen2020simple}, which utilized a siamese architecture in conjunction with an InfoNCE~\cite{oord2018representation} contrastive loss. MoCo~\cite{he2020momentum} implemented a memory bank to store negative samples, but several studies have shown that negative pairs are not always necessary. BYOL~\cite{richemond2020byol} employed an asymmetric architecture with EMA, where the online network attempts to predict the representations of the target network. SimSiam~\cite{chen2021exploring} provided a straightforward siamese framework that incorporated a stop-gradient operation on one branch and an additional predictor step on the other. More recently, SSL has begun to adopt the use of vision transformers~\cite{dosovitskiy2020image}. DINO~\cite{caron2021emerging} combined a distillation loss with vision transformers in a siamese framework. 

\section{Approach}

In this section, we first introduce each component of our framework elaborately, including: {(1)} a filling-based masking strategy; {(2)} an asymmetric loss formulation with soft distance to match the proposed mix-masking scheme. Then, we provide an overview of the proposed architecture comparing to the basic model and differences from other counterparts.

\subsection{Masking Strategy}
\vspace{-0.1in}

\begin{figure}[!htb]
  \centering
  \vspace{-0.15in}
  \includegraphics[width=0.8\textwidth]{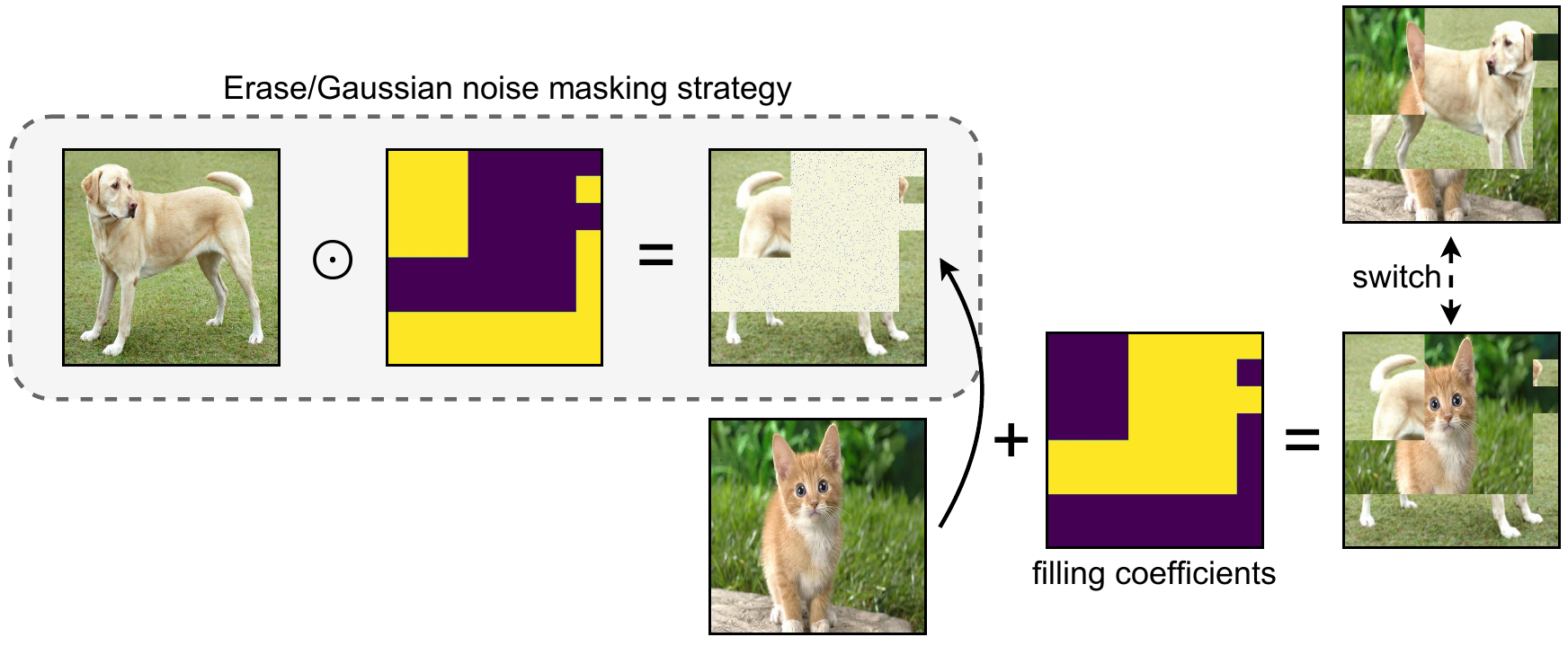}
  \caption{Illustration of the proposed filling-based masking strategy. The gray dashed box shows Erase/Gaussian noise \cite{jing2022masked} masking strategy. A formal definition of a switch image in the case of reverse permutation is given in Eq. \ref{eq:switch}.}
  \label{fig:masking}
\end{figure}

\begin{figure*}[t]
  \centering
  \includegraphics[width=0.9\textwidth]{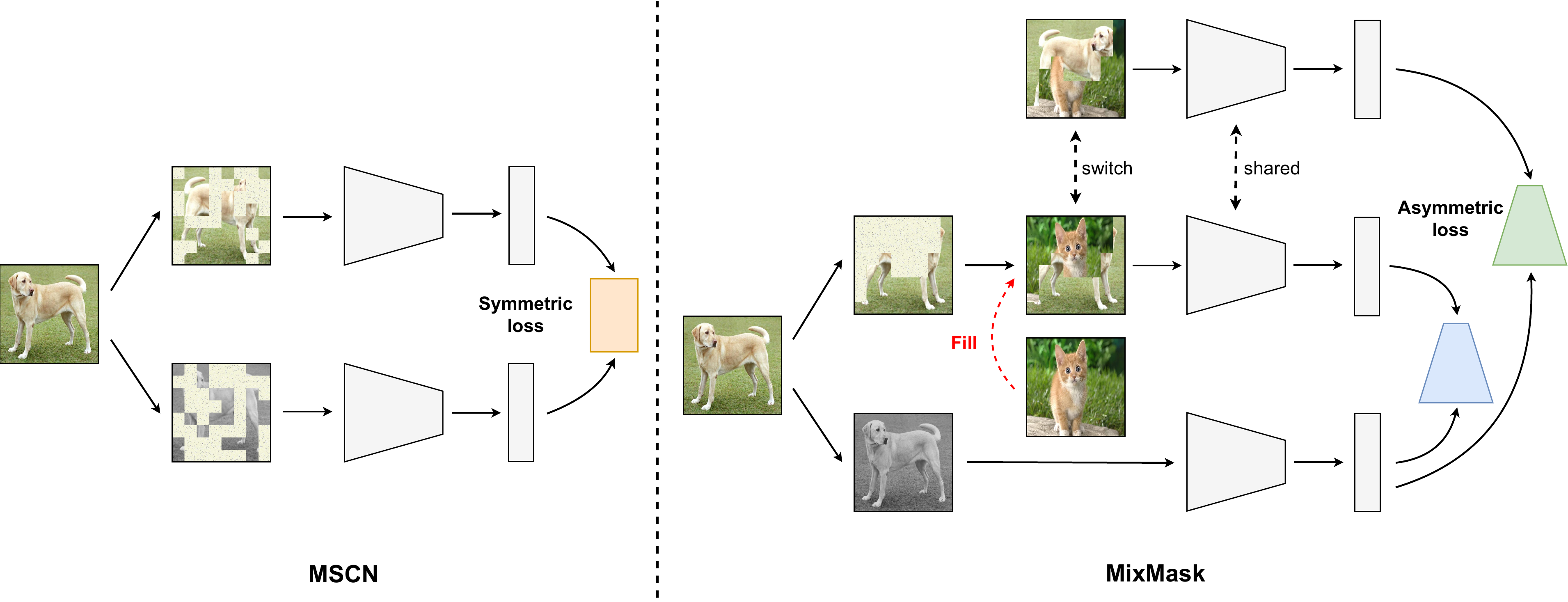}
  \vspace{-0.1in}
  \caption{Illustration of the Masked Siamese ConvNets (left) and our proposed framework (right). MixMask branch incorporates asymmetry into the loss function design by generating images with different rates of similarity to the images in the original branch. In MixMask branch image of the truck is presented twice with different levels of similarity to the image in the original branch due to the regions masked with contents of another image.}
  \label{fig:overview}
\end{figure*}

\noindent{\bf Masking Scheme: Erasing or Filling?} In Fig.~\ref{fig:masking}, the prior {\em Erase-based} {\em Gaussian noise masking} strategy is illustrated inside the dashed box. In such strategy, erased regions can be filled with a Gaussian noise \cite{jing2022masked}. Different from Masked Image Modeling (MIM), which is to reconstruct the masked contents for learning good representation, the {\em Masked Siamese Networks} will not predict the information in removed areas, so erasing will only lose information and is not desired in the learning procedure. In contrast to erase-based masking, our filling-based strategy will repatch the removed areas using an auxiliary image, as shown in the right part of Fig.~\ref{fig:masking}. After that, we will {\em switch} the content between the main and auxiliary images to generate a new image for information completeness of the two input images.  For a given original image $\bm I_i$ we define its mixture as a mix of the pair {$(\bm I_i, \bm I_{n - i})$} and \textit{switch} image of $\bm I_i$ as a mixture of the pair $(\bm I_{n - i}, \bm I_i)$. Although the mixture contains information from another image, which may have different semantics, it will still be beneficial by cooperating with the soft loss design, discussed in Section \ref{section:loss_design}, which takes into account the degree of each individual image included in the mixture.

%Our soft objective calculation is designed to fit the format of such masked images in siamese networks, as described in the next section. 

\noindent{\bf Masking Pattern: Blocked or Discrete?}
Masking pattern determines the difficulty for the model to generate representations in siamese networks, even if the masking ratio is the same, the representation will still be different for different masking patterns. It will directly influence the information of input to further affect the latent representations.
{\em Our observation on Masked Siamese ConvNets is opposite to that in MIM methods, which found discrete/random masking is better}~\cite{he2022masked, xie2022simmim}. From our empirical experiments, on CIFAR-100, blocked and discrete masking patterns achieve similar accuracy, and discrete is slightly better, however, on Tiny-ImageNet and ImageNet-1K, blocked mask clearly shows superiority over discrete/random. We explain this as that if the input size is small, the mask pattern is not so important since the semantic information of the object is still preserved. On larger datasets like Tiny-ImageNet and ImageNet-1K, discrete/random masking will entirely destroy the completeness of the object in an image, as shown in Fig.~\ref{fig:masking_all} (d, e),  while this is crucial for ConvNet to extract a meaningful representation of the object. {Blocked masking also better synergizes with the contrastive objective, which is biased to learn global features~\cite{chen2021intriguing, cole2022does}}. Therefore, blocked masking shows superior ability on the MSCN and is a better choice than random discrete masking.

\subsection{Distance in Siamese Networks}
\label{section:loss_design}
\noindent{\bf{Objective Calculation: Inflexible or Soft?}} It has been observed~\cite{shen2022unmix} that different pretext and data processes (e.g., masking, mixture) will change the semantic distance of two branches in the siamese networks, hence the default symmetric loss will no longer be aligned to reflect the true similarity of the representations. It thus far has not attracted enough attention for such a problem in this area. In this work, we introduce a soft objective calculation method that can fit the filling-based masking strategy in a better way.  
To calculate the soft distance, we start by generating a binary mask with a fixed grid size that will later be used to mix a batch of images, denoted as $\bm I$, from a single branch. {In case when we use a reverse permutation to obtain the mixture}, each image in the batch with index $i$ is mixed according to the mask with the image in the same batch but with index $n-i$ as described in Eq. \ref{eq:mixture}:
\small 
\begin{flalign}
 \vspace{-0.2in}
 \text{mix}_i &= \text{mix}(\bm I_{i}, \bm I_{n - i}) = m \odot \bm I_{i} + (1 - m) \odot \bm I_{n - i} \label{eq:mixture} \\
\text{switch}_i &= \text{mix}(\bm I_{n - i}, \bm I_{i}) = m \odot \bm I_{n -i} + (1 - m) \odot \bm I_{i} \label{eq:switch}, 
\end{flalign}\normalsize
where $\bm n$ is batch size and $m$ is mask. The mixed image contains parts from both $\bm I_{i}$ and $\bm I_{n - i}$ whose spatial locations in the mixture are defined by the contents of the binary mask. {Using $m$ and $1 - m$ ensures that each region in the mixture will contain pixels from exactly one image and that there will be no empty unfilled regions.}
Furthermore, {to reflect the contribution of each image in the mixture,} we calculate a mixture coefficient $\lambda$, which is equal to the ratio of the masked area to the total area of the image using the Eq. \ref{eq:lambda}:
\begin{equation}
\label{eq:lambda}
\small
\lambda = \frac{\sum \limits_{x, y} \mathds{1}[mask(x, y)=1]}{{width} \cdot {height}},   
\end{equation}
where $\mathds{1}$ is the indicator function that measures the masked area of an image.\\ 
\textbf{Loss Function.} The final loss is defined as a summation of the original loss and mixture loss:
$
\mathds{L} = \mathds{L}_\text{Orig} + \mathds{L}_\text{MixMask}. $
Here we take a contrastive loss from MoCo \cite{he2020momentum} as an example, {where $q$ and $k$ are two randomly augmented non-mixed views of the input batch. The notation stems from queries and keys from MoCo nomenclature}, the $\mathds L_\text{Orig}$ will be:
$\mathds L_\text{Orig}=-  \log \frac{\exp \left(q \cdot k / \tau\right)}{\sum_{i=0}^{K} \exp \left(q \cdot k_{i} / \tau\right)}$.
The mixture term $\mathds L_\text{MixMask}$ will contain two terms which are scaled with $\lambda$ coefficient:
\begin{equation}
\label{eq:mixmask_loss}
\small
\mathds{L}_{\text{\tiny MixMask}} = \lambda \mathds{L}_{\uparrow} + (1 - \lambda) \mathds{L}_{\downarrow} = -\bigg(\lambda \log \frac{\exp (q_{\uparrow} \cdot k / \tau)}{\sum_{i=0}^{K} \exp (q_{\uparrow} \cdot k_i / \tau)} + (1-\lambda) \log \frac{\exp (q_{\downarrow} \cdot k / \tau)}{\sum_{i=0}^{K} \exp (q_{\downarrow} \cdot k_i / \tau)}\bigg),
\end{equation}

where $q_{\uparrow}$ and $q_{\downarrow}$ are normal and reverse orders of mixed queries in a mini-batch, $k$ is the unmixed single key, $\lambda$ is calculated using the Eq. \ref{eq:lambda} and $\tau$ is the temperature.

\subsection{Framework Overview}

Our framework overview is shown in Fig.~\ref{fig:overview}. In this figure, the left is the conventional  Masked Siamese ConvNets (MSCN), right is our proposed MixMask with asymmetric distance loss. The motivation behind this design is that directly erasing regions will lose a significant proportion of information in the Siamese ConvNets, which cannot be recovered by post-training. This is quite different from the mechanism of Masked Autoencoders (MAE) \cite{he2022masked} that predict masked areas to learn good representations. According to this, we propose a filling-based scheme to overcome the drawback. The soft distance loss is designed to fit the true similarity of the two branches. We empirically show that with the integrality of mix-masking and objective, we can learn more robust and generalized representations from the masked input.

\noindent{\bf Differences from Prior Counterparts.} 
MSN and MSCN utilize regular erase-based masking and Un-Mix approach integrates Mixup and CutMix techniques into siamese networks for self-supervised learning. In contrast, our proposed MixMask method is a generalized masking approach specifically designed for siamese networks that allows for arbitrary mask area shapes. 
 
Our empirical study demonstrates that MixMask exhibits stronger representation learning capabilities. Interestingly, MixMask is also compatible with Un-Mix~\cite{shen2022unmix} that can be employed jointly to further improve performance and achieve state-of-the-art accuracy. {SparK and ConvNeXt V2 propose a way of doing MIM with single branch ConvNet and reconstruction loss, while our work is focused on the Siamese contrastive framework.}

\section{Experiments}

\noindent{\textbf{Base Models.}} 
In our experimental section, we use base models including: { MoCo V1\&V2~\cite{he2020momentum, chen2020improved}}, { Un-Mix~\cite{shen2022unmix}}, { SimCLR~\cite{chen2020simple}}, { BYOL~\cite{richemond2020byol}} and { SimSiam~\cite{chen2021exploring}}.

\noindent{{\bf Datasets and Training Settings.}} 
We conduct experiments on {Tiny-ImageNet~\cite{le2015tiny}}, {CIFAR-100~\cite{krizhevsky2009learning}},   and {ImageNet-1K~\cite{deng2009imagenet}}. For CIFAR-100 and Tiny-ImageNet, we train each framework for 1000 epochs with ResNet-18 \cite{he2016deep}. For ImageNet-1K, we pretrain ResNet-50 for 200 epochs and then finetune a linear classifier on top of the frozen features for 100 epochs and report Top-1 accuracy. We use MoCo (and MoCo V2 for ImageNet-1K) as a base framework unless stated otherwise. We report Top-1 on linear evaluation, except the case of MoCo on CIFAR-100 and Tiny-ImageNet for which we provide {\em k}-NN accuracy. $^*$ indicates that we build our method upon Un-Mix. All our experiments are conducted on A100 (40G) GPUs.

\subsection{Ablation Study}
In the experiments, we use a blocked masking strategy, masking ratio of 0.5, and grid size of 2, 4, 8 for CIFAR-100, Tiny-ImageNet, and ImageNet-1K if not stated otherwise.

\noindent{{\bf Masking Strategy.}}  
We first explore how different masking strategies affect the final result. We consider three different hyperparameters: grid size, grid strategy, and masking ratio. To make our experiments more reliable, we consider three datasets with different image sizes: CIFAR-100: 32$\times$32, Tiny-ImageNet: 64$\times$64, and ImageNet-1K: 224$\times$224. 

\begin{wraptable}{r}{0.48\textwidth}
\vspace{-0.2in}
\begin{center}
\resizebox{0.48\textwidth}{!}{
\begin{tabular}{cccc}
\multicolumn{1}{c}{}  &\multicolumn{1}{c}{\bf CIFAR-100} & \multicolumn{1}{c}{\bf Tiny-ImageNet} & \multicolumn{1}{c}{\bf ImageNet-1K}\\  \hline
\bf Input size   & 32$\times$32   & 64$\times$64 & 224$\times$224 \\ \hline
\multicolumn{1}{c}{\bf Grid Size} & \multicolumn{3}{c}{} \\  
2$\times$2   & \cc \bf 68.1  & 46.4 & 68.4 \\
4$\times$4   & 67.6  & \cc \bf 47.4 & 69.0 \\
8$\times$8   & 67.5  & 46.5 & \cc \bf 69.2 \\  
16$\times$16   & --  & 46.1 & 68.5 \\  
32$\times$32   & --  & -- & 68.7 \\  
48$\times$48   & --  & -- & 68.8 \\  \hline
\multicolumn{1}{c}{\bf Masking Strategy} & \multicolumn{3}{c}{} \\  
blocked  & 67.8 & \cc \bf 47.4 & \cc \bf 69.2 \\
discrete  & \cc \bf 68.1 & 46.4 & 68.4 \\  \hline
\multicolumn{1}{c}{\bf Masking Ratio} & \multicolumn{3}{c}{} \\ 
0.25$|$0.75     & 66.8 & 45.6 & 68.5 \\
0.5   & \cc \bf 68.1  & \cc \bf 47.4 & \cc \bf 69.2 \\
uniform(0, 1)  & 67.6 & 45.5 & 68.7 \\ \hline
%uniform(0.25, 0.75) & 67.8 & 47.0 & 68.9 \\ \hline
\multicolumn{1}{c}{\bf Switch Mixture} & \multicolumn{3}{c}{} \\  
Yes  & 
\bf \cc 68.1  & \cc \bf 47.4 & \cc \bf 69.2 \\
No  & 67.3  & 45.6 & 67.8 \\  
\end{tabular}
}
\caption{Ablation study on masking and {switch} strategies using MoCo V1/V2 with original and MixMask branches. {\em k}-NN accuracy averaged over 3 runs is reported.}
\vspace{-0.4in}
\label{table:mask_strategies}
\end{center}
\end{wraptable}
The first parameter grid size specifies the granularity of the $n\times n$ square grid of the image. We consider the following values of $n=2,4,8,16,32, 48$. However, we have different upper bounds for different datasets depending on the spatial size of the images. From our experiments, we can conclude that a very large grid size completely disrupts the semantic features of the image and leads to poor performance. Our results indicate that optimal grid size increases proportionally to the input size of the image. We obtain the optimal grid size for CIFAR-100 is 2, for Tiny-ImageNet is 4, and for ImageNet-1K is 8. The masks with very small and very large grid sizes show bad performance. We think this happens because a small grid size does not provide enough variance in the mask structure whilst a large grid size destroys the important semantic features of the image.

We consider two different strategies for the random mask generation, discrete/random mask and blocked mask. We generate a blocked mask according to the algorithm described in \cite{bao2021beit}. A discrete mask does not have any underlying structure, whilst a blocked mask is generated in a way to preserve global spatial continuity and, thus, is more suitable for capturing global semantic features. For CIFAR-100, we observe a negligible difference between blocked and discrete masks. On the other hand, for Tiny-ImageNet and ImageNet-1K, which have larger spatial sizes of the image, blocked mask performs better than discrete. This observation is different from {MIM-based approaches \cite{he2022masked, xie2022simmim}}, where a discrete mask achieves better performance and highlights the importance of maintaining global features when generating an image mixture as opposed to the case when erasing parts of the image by performing a vanilla masking operation. This also reflects the different mechanisms and requirements of reconstruction-based and siamese-based self-supervised learning approaches. 

For the masking ratio, we consider two cases with constant values of $0.5$ and $0.25/0.75$ (as there is no difference between these two due to the loss function design) and two cases when the value is sampled from the uniform distribution with different bounds. We obtain the best results for masking ratio $0.5$ under different settings, conjecturing that this value causes the blocked mask to generate consistent global views for both of the images being mixed. Sampling masking ratio from a uniform distribution with bounds $0.25$ and $0.75$ yields better results than using $0$ and $1$ as bounds. We believe this shows that extreme values close to either $0$ or $1$ generate a mixture where one image heavily dominates over the other when in the optimal mixture, areas of each image should be roughly proportional.

\noindent{\textbf{Switch Mixture.}} 
{We also examine the efficacy of the second term in the mixture loss in Eq.~\ref{eq:mixmask_loss}, which is computed using switch images and multiplied with soft coefficient $(1 - \lambda)$. Certainly, having two terms in the mixture loss part is beneficial and gives better results in Table~\ref{table:mask_strategies} on all datasets as it provides more training signal.}

\begin{figure*}[!ht]
  \centering
  \begin{minipage}{0.99\textwidth}
    \includegraphics[width=\textwidth]{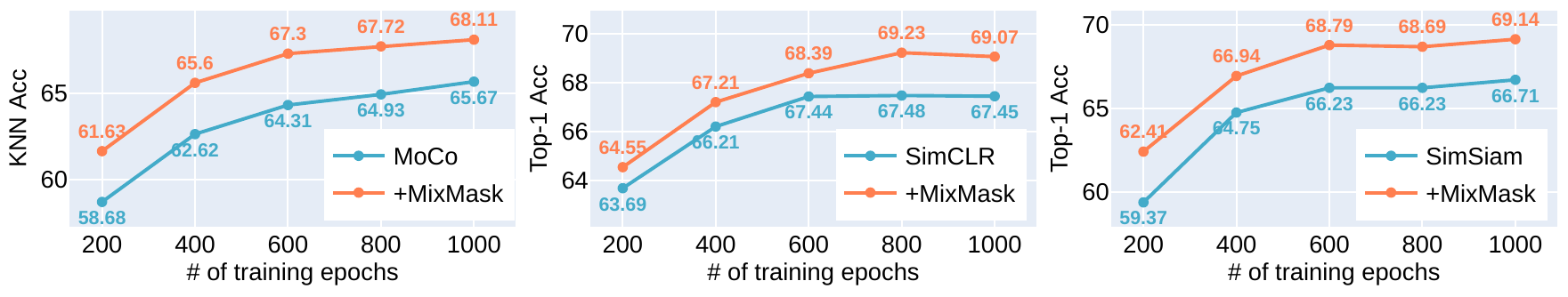}
    \caption{Results with different training budgets and base frameworks on CIFAR-100. MixMask consistently performs better than the baseline for every configuration.}
    \label{fig:training_budget}
  \end{minipage}
\end{figure*}

\noindent{{\bf Training Budgets.}} 
We test our method with different training budgets of 200, 400, 600, 800, and 1000 epochs on CIFAR-100 using MoCo, SimCLR, and SimSiam. The results are shown in Fig.~\ref{fig:training_budget}. We can observe our method achieves consistent improvement over various frameworks.

\noindent{{\bf Results for Different Base Frameworks.}}
In Table \ref{table:framework} we consider the generalizability of our method by applying it on top of four different self-supervised learning frameworks. 
When applying our method upon Un-Mix, we yield superior performance in all cases. {Plain MixMask also provides a  competitive performance in all the experiments.}

\begin{table}[!t]
\begin{center}
 \resizebox{0.795\textwidth}{!}{
\begin{tabular}{lcccccccc}
\multicolumn{1}{c}{}  &\multicolumn{4}{c}{\bf CIFAR-100} & \multicolumn{2}{c}{\bf Tiny-INet}  & \multicolumn{1}{c}{\bf  INet-1K}\\  \hline
& \multicolumn{1}{c}{\bf MoCo} &  \multicolumn{1}{c}{\bf SimSiam} &  \multicolumn{1}{c}{\bf BYOL} &  \multicolumn{1}{c}{\bf SimCLR} & \multicolumn{1}{c}{\bf MoCo}& \multicolumn{1}{c}{\bf BYOL} & \multicolumn{1}{c}{\bf MoCoV2} & 
\\  \hline
Vanilla   & 65.7  & 66.7 & 66.8 & 67.1 & 42.8 & 51.4 & 67.5 \\ 
Un-Mix~\cite{shen2022unmix} & 68.6 & 70.0 & 70.0 & 69.8 & 45.3 & 52.9 & 68.5  \\ 
MixMask & 68.4 & 69.8 & 71.9 & 69.2 & 47.4 & 50.8 & 69.2 \\
MixMask$^*$  & \cc \bf 70.0 & \cc \bf 70.6 & \cc \bf 72.1 & \cc \bf 70.0 & \cc \bf 47.9 & \cc \bf 53.7 & \cc \bf 69.5 \\
\end{tabular}
}
\caption{Ablation on different base frameworks. * means we use our method with Un-Mix.}
\vspace{-0.3in}
\label{table:framework}
\end{center}
\end{table}

\begin{table*}[!ht]
\begin{minipage}{0.55\textwidth}
\resizebox{\textwidth}{!}{
\begin{tabular}{cccccccc}
Method & Top-1 200 ep & COCO AP\textsubscript{det} & COCO AP\textsubscript{seg} & Top-1 1\% labels  \\ \hline
MoCo V2 & 67.5 & 39.3  &  34.4 & 50.3  \\
MSCN  & 68.2 & 39.1 & 34.2 & 54.0  \\ 
MixMask & \cc \textbf{69.2} & \cc \textbf{39.8} & \cc \textbf{34.8} & \cc \textbf{54.5} 
\end{tabular}
}
\vspace{0.1in}
\caption{MixMask shows better performance than MoCo v2 on linear probing, COCO detection and segmentation, and 1\% label supervised finetuning.}
\label{table:mscn_comparison}
\end{minipage}
\hfill
\begin{minipage}{0.44\textwidth}
\includegraphics[width=0.95 \textwidth]{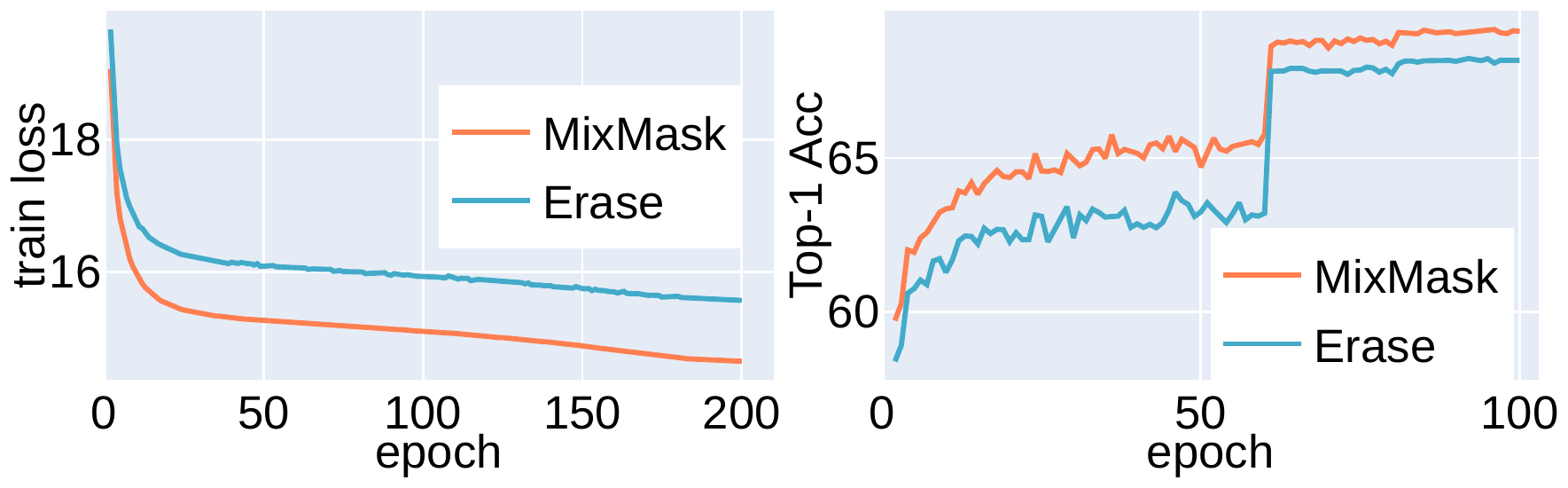}
\captionof{figure}{MixMask outperforms MSCN on ImageNet-1K by 1\%.}
\label{fig:plots}
\end{minipage}
\end{table*}

\begin{table*}[!ht]
\begin{minipage}[c]{0.35\textwidth}
\begin{center}
\resizebox{0.99\textwidth}{!}{
\begin{tabular}{lccc}
\multirow{2}{*}{Method} & \multicolumn{1}{c}{\bf 1\% labels} & \multicolumn{1}{c}{\bf 10\% labels} & \multicolumn{1}{c}{\bf 100\% labels} \\
\cline{2-4}
  & Top-1 Acc & Top-1 Acc & Top-1 Acc \\ \hline
MoCo V2~\cite{chen2020improved} & 50.3 & 66.8 & 76.7 \\
SwAV~\cite{caron2020unsupervised} & 53.9 & \bf 70.2 & -- \\
Barlow Twins~\cite{zbontar2021barlow} & \bf 55.0 & 69.7 & -- \\ \hline
MSCN~\cite{jing2022masked}   & 54.0 & \-- & --  \\
MixMask & \cc \bf 54.5 &\cc 69.3 & \cc \bf 77.7  
\end{tabular}}
\caption{{Results of semi-supervised and supervised finetuning on ImageNet-1K.}}
\label{table:semi}
\end{center}
\end{minipage}
\hfill
\begin{minipage}[c]{0.64\textwidth}
\begin{center}
 \resizebox{0.99\textwidth}{!}{
\begin{tabular}{lccccccccc}
\multicolumn{1}{c}{\bf } &  \multicolumn{6}{c}{\bf MS COCO 2017} & \multicolumn{3}{c}{\bf Pascal VOC 2007} \\  
\cline{1-10}
\multirow{2}{*}{Method} & \multicolumn{3}{c}{\bf Object detection} & \multicolumn{3}{c}{\bf Segmentation} & \multicolumn{3}{c}{\bf Object detection} \\
\cline{2-10}
  & AP & AP$_{50}$ & AP$_{75}$ & AP & AP$_{50}$ & AP$_{75}$ & AP & AP$_{50}$ & AP$_{75}$\\ \hline
MoCo V2~\cite{chen2020improved} & 39.3 & 58.9 & 42.5 & 34.4 & 55.8 & 36.5 & 57.4 & 82.5 & 64.0\\
SwAV~\cite{caron2020unsupervised} & 38.4 & 58.6 & 41.3 & 33.8 & 55.2 & 35.9 & 56.1 & 82.6 & 62.7  \\
Barlow Twins~\cite{zbontar2021barlow} & 39.2 & 59.0 & 42.5 & 34.3 & 56.0 & 36.5 & 56.8 & 82.6 & 63.4  \\ \hline
MSCN~\cite{jing2022masked} & 39.1 & 59.1 & 42.1 & 34.2 & 55.7 & 36.4 & 57.5 & \bf 83.0 & 64.4  \\
MixMask & \cc \bf 39.8 & \cc \bf 60.0 & \cc \bf 43.0 & \cc \bf 34.8 & \cc \bf 56.3 & \cc \bf 37.0 & \cc \bf 57.9 & \cc \bf 83.0 & \cc \bf 64.5
\end{tabular}
}
\caption{MixMask gives the best performance on all metrics on COCO and Pascal VOC.}
\label{table:det_seg}
\end{center}
\end{minipage}
\end{table*}

\subsection{Comparison and Superiority to MSCN}
 We present evidence demonstrating that our approach, MixMask, outperforms MSCN \cite{jing2022masked}. Specifically, Table~\ref{table:mscn_comparison} shows the results of linear probing for 200 epoch pretraining and downstream evaluation for object detection/segmentation on COCO~\cite{lin2014microsoft} and supervised fine-tuning on 1\% of ImageNet-1K. MixMask achieves superior performance by proposing a method to fill the erased regions with meaningful semantics on all of the benchmarks. Additional results for semi-supervised fine-tuning and object detection are available in Table~\ref{table:det_seg} and~Table~\ref{table:semi}. The results demonstrate that MixMask outperforms MSCN and even surpasses the performance of SwAV in semi-supervised fine-tuning. Moreover, MixMask achieves better results than MSCN on all evaluation metrics for detection and segmentation. Notably, our approach is conceptually simpler and computationally more efficient than MSCN, which requires combining multiple masking strategies, adding Gaussian noise, and multicrops. 
 
 \vspace{-0.08in}
\subsection{Results on Semi- and Supervised Finetuning} \label{sec:semisup}
For the semi and supervised finetuning on ImageNet-1K, we follow the protocol described in \cite{jing2022masked}. We explore three different data regimes of 1\%, 10\%, and 100\% of labels and finetune all models for 20 epochs. The results are given in Table~\ref{table:semi}, MixMask outperforms MSCN as well as MoCoV2 and SwAV on 1\% data regime and shows competitive performance to SwAV and Barlow Twins on 10\%.

\vspace{-0.1in}
\subsection{Results on Object Detection and Segmentation}
\label{sec:detection}
We test our method on the downstream task of object detection and segmentation. For that, we finetune a Faster-RCNN \cite{ren2015faster}, and Mask-RCNN \cite{he2017mask} models from Detectron2 \cite{wu2019detectron2} on Pascal VOC 2007 \cite{everingham2010pascal}, and MS COCO 2017 \cite{lin2014microsoft} datasets. For VOC 2007, we follow the standard evaluation protocol in \cite{he2020momentum} with 24k training iterations. For COCO, we use $1\times$ schedule from \cite{wu2019detectron2}. {The results in Table~\ref{table:det_seg} verify the superiority of the proposed MixMask to all other frameworks, including MSCN.} 

\section{Conclusion}

We have introduced MixMask, a new masking approach tailored for self-supervised learning on ConvNets. This method harmoniously merges an asymmetric loss with a filling-based masking procedure within the framework of the Masked Siamese ConvNets. Our experiments consistently demonstrate that MixMask surpasses the conventional symmetric loss and erase-centric masking techniques used in MSCN. We have demonstrated that our distinct asymmetric loss function outshines the standard symmetric loss associated with MSCN. Moreover, we addressed the challenge of diminished learning efficiency inherent in traditional erase-centric masking by adopting a filling-centric approach. In this strategy, an alternative image serves to occupy the voided regions. We have also laid out a comprehensive guideline for seamlessly infusing MixMask into other blending techniques, elaborating on how varying masking parameters influence the caliber of the resultant representations. The results unequivocally highlight our method's superiority over MSCN in linear probing and multiple downstream applications such as object detection and segmentation.

\bibliography{egbib}
\newpage 
\section*{Appendix}

\appendix

\section{Base Models \& Datasets}

In this section, we provide a description of self-supervised learning frameworks and datasets that we used in the experiments. To test our method, we tried to select a diverse set of frameworks that incorporate different mechanisms to avoid model collapse and follow different design paradigms.
\subsection{Base Models}
\noindent{\bf MoCo V1\&V2~\cite{he2020momentum, chen2020improved}} is a self-supervised contrastive learning framework that employs a memory bank to store negative samples. MoCo V2 is an extension of the original MoCo, which introduces a projection head and stronger data augmentations.

\noindent{\bf Un-Mix~\cite{shen2022unmix}} is an image mixture technique with state-of-the-art performance for unsupervised learning, which uses CutMix and Mixup at its core. It smooths decision boundary and reduces overconfidence in model predictions by introducing an additional mixture term to the original loss value, which is proportional to the degree of the mixture. 

\noindent{\bf SimCLR~\cite{chen2020simple}} 
is a siamese framework with two branches that uses contrastive loss to attract positive and repel negative instances using various data augmentations.

\noindent{\bf BYOL~\cite{richemond2020byol}} 
 is a self-supervised learning technique that does not use negative pairs. It is composed of two networks, an online and a target. The task of an online network is to predict the representations produced by the target network. EMA from the online network is used to update the weights of the target. 

\noindent{\bf SimSiam~\cite{chen2021exploring}} 
The authors examined the effect of the different techniques which are commonly used to design siamese frameworks for representation learning. As a result, they proposed a simple framework with two branches that relies on the stop gradient operation on one branch and an extra prediction module on the other.

\subsection{Datasets}

\noindent{\bf CIFAR-100~\cite{krizhevsky2009learning}} 
consists of 32$\times$32 images with 100 classes. There are 50,000 train images and 10,000 test images, 500 and 100 per class, respectively.

\noindent{\bf Tiny-ImageNet~\cite{le2015tiny}} 
is a dataset containing 64 $\times$ 64 colored natural images with 200 classes. The test set is composed of 10,000 test images, whilst the train contains 500 images per category, totaling 100,000 images. 

\noindent{\bf ImageNet-1K~\cite{deng2009imagenet}} 
has images with a size of 224$\times$224. 1,281,167 images span the training set, with 1K different classes, and the validation set includes 50K images.

\section{Training Configurations}

In this section we provide hyperparameter settings for: 
\begin{itemize}
\itemsep0em
\item Training on CIFAR-100 and Tiny-ImageNet in Table \ref{hparams_cifar}.
\item Pretraining and linear probing on ImageNet-1K configurations are shown in Table \ref{hparams_inet}.
\item Configurations for semi-supervised and supervised fine-tuning on ImageNet-1K are given in Table \ref{hparams_inet_ft}.
\item For object detection and segmentation we use the Detectron2\cite{wu2019detectron2} library and follow the $1\times$ recipe on COCO and standard 24k training protocol on VOC07.
\end{itemize}

\begin{table*}[h]
\centering
\begin{tabular}{cc |cc|cc}
\multicolumn{2}{c|}{MoCo} & \multicolumn{2}{c|}{SimCLR \& BYOL} & \multicolumn{2}{c}{SimSiam} \\
\hline
hparam & value & hparam & value & hparam & value \\ \shline
backbone & resnet18 & backbone & resnet18 & backbone & resnet18 \\
 optimizer & SGD &  optimizer & Adam & optimizer & SGD \\
 lr & 0.06 & lr & 0.003\slash0.002 &  lr &  0.03\\  
 batch size & 512 & batch size & 512 & batch size & 512 \\ 
opt momentum & 0.90  & proj layers & 2 & opt momentum & 0.90  \\ 
 epochs & 1,000 & epochs & 1,000 &  epochs & 1,000\\
 weight decay & 5e-4 & weight decay & 5e-4 &  weight decay & 5e-4 \\
 embed-dim & 128 & embed-dim & 64\slash 128 & embed-dim & 128\\
 moco-m & 0.99 & Adam l2 & 1e-6 &  warmup epochs & 10 \\
 moco-k & 4,096 &  proj dim & 1,024 &  proj layers & 2\\
 unmix prob & 0.50 &  unmix prob & 0.50 & unmix prob & 0.50\\
 moco-t & 0.10 & byol tau & 0.99 &  & \\
\end{tabular}
\caption{Training settings on CIFAR-100 and Tiny-ImageNet. Slash separated values correspond to CIFAR-100 and Tiny-ImageNet, respectively. }
\label{hparams_cifar}
\end{table*}
\begin{table*}[h]
\begin{minipage}[c]{0.48\textwidth}
\resizebox{0.999\textwidth}{!}{
\begin{tabular}{cc|cc}
\multicolumn{2}{c}{Pretraining} & \multicolumn{2}{|c}{Linear probing} \\ \hline
hparam & value & hparam & value  \\ \shline
backbone & resnet50 & backbone & resnet50  \\
optimizer & SGD & optimizer & SGD\\
lr & 0.03 & lr & 30\\
batch size & 256 & batch size & 256\\
opt momentum & 0.90 & opt momentum & 0.90\\
lr schedule & cosine & lr schedule & [60, 80]\\
epochs & 200/800  & epochs & 100\\
weight decay & 0 & weight decay & 0 \\
moco-dim & 128 \\
moco-m & 0.999 \\
moco-k & 65,536 \\
moco-t & 0.2 \\
unmix probability & 0.5 \\
mask type & block \\ 
grid size & 8 \\
\end{tabular}}
\caption{Hyperparameter values for pretraining and linear probing on ImageNet-1K. This configuration achieves the highest score. All experiments are conducted on 4 $\times$ NVIDIA A100 SXM4 40GB
 GPU.}
\label{hparams_inet}
\end{minipage}
\hfill
\begin{minipage}[c]{0.48\textwidth}
\begin{center}
\begin{tabular}{cc}
hparam & value \\ \shline
backbone & resnet50 \\
 optimizer & SGD \\
 lr stem & 0.002\slash0.002\slash0.001 \\
 lr classifier & 0.5\slash 0.5\slash 0.05 \\
 batch size & 256 \\
opt momentum & 0.90 \\
lr schedule & [12, 16] \\
 epochs & 20 \\
 weight decay & 0
\end{tabular}
\caption{Hyperparameter values for semi-supervised and supervised finetuning on ImageNet-1K. Slash separated values correspond to 1\%, 10\% and 100\% percent data regimes, respectively.}
\label{hparams_inet_ft}
\end{center}
\end{minipage}
\end{table*}

\begin{table*}

\hfill
\begin{minipage}[c]{0.48\textwidth}

\begin{center}

\end{center}
\end{minipage}
\vspace{-0.2in}
\end{table*}

% \newpage

\section{Training Loss and Accuracy Curves}

\begin{figure*}[!htb]
  \centering
  \includegraphics[width=0.93\textwidth]{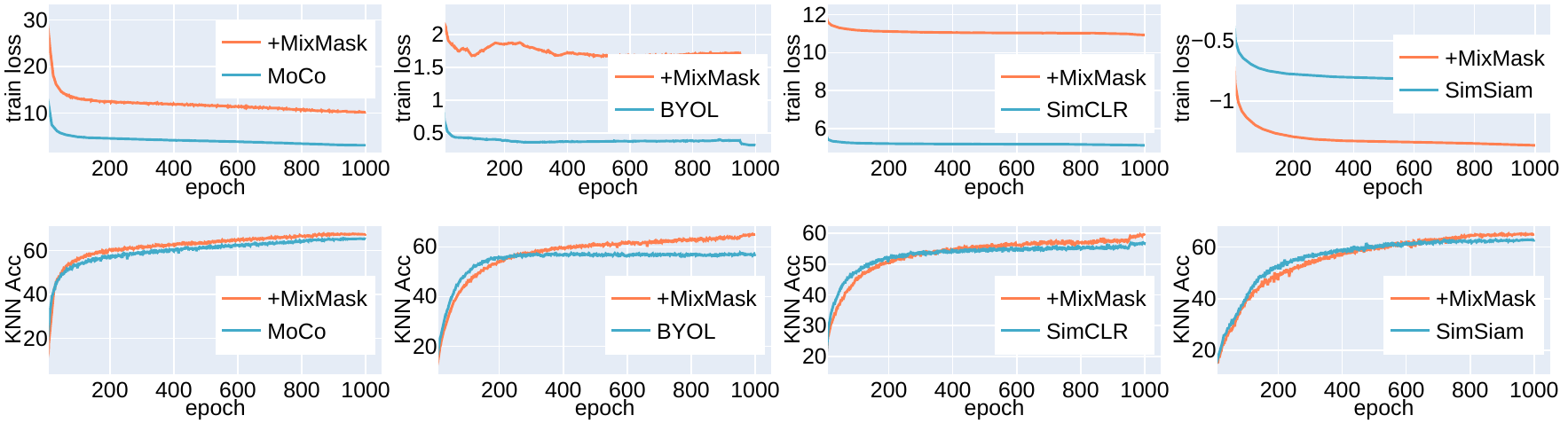}
  \caption{Training losses (top row) and {\em k}-NN evaluation accuracies (bottom row) on CIFAR-100 for experiments with 1,000 epochs for different self-supervised frameworks. MixMask (red) outperforms vanilla baseline (blue) on all frameworks by a significant margin.}
  \label{fig:plots_all}
\end{figure*}

In Fig. \ref{fig:plots_all}, we present the training loss and {\em k}-NN accuracy curves for different base frameworks trained for 1,000 epochs on CIFAR-100 dataset. MixMask consistently outperforms baseline on all methods. MixMask has a higher (in case of SimSiam lower because it can attain the value of -1) training loss than baseline due to the presence of the additional asymmetric loss term.

\section{Illustrations of Different Mask Patterns}
\begin{figure*}[!htb]
  \centering
  \includegraphics[width=0.87\textwidth]{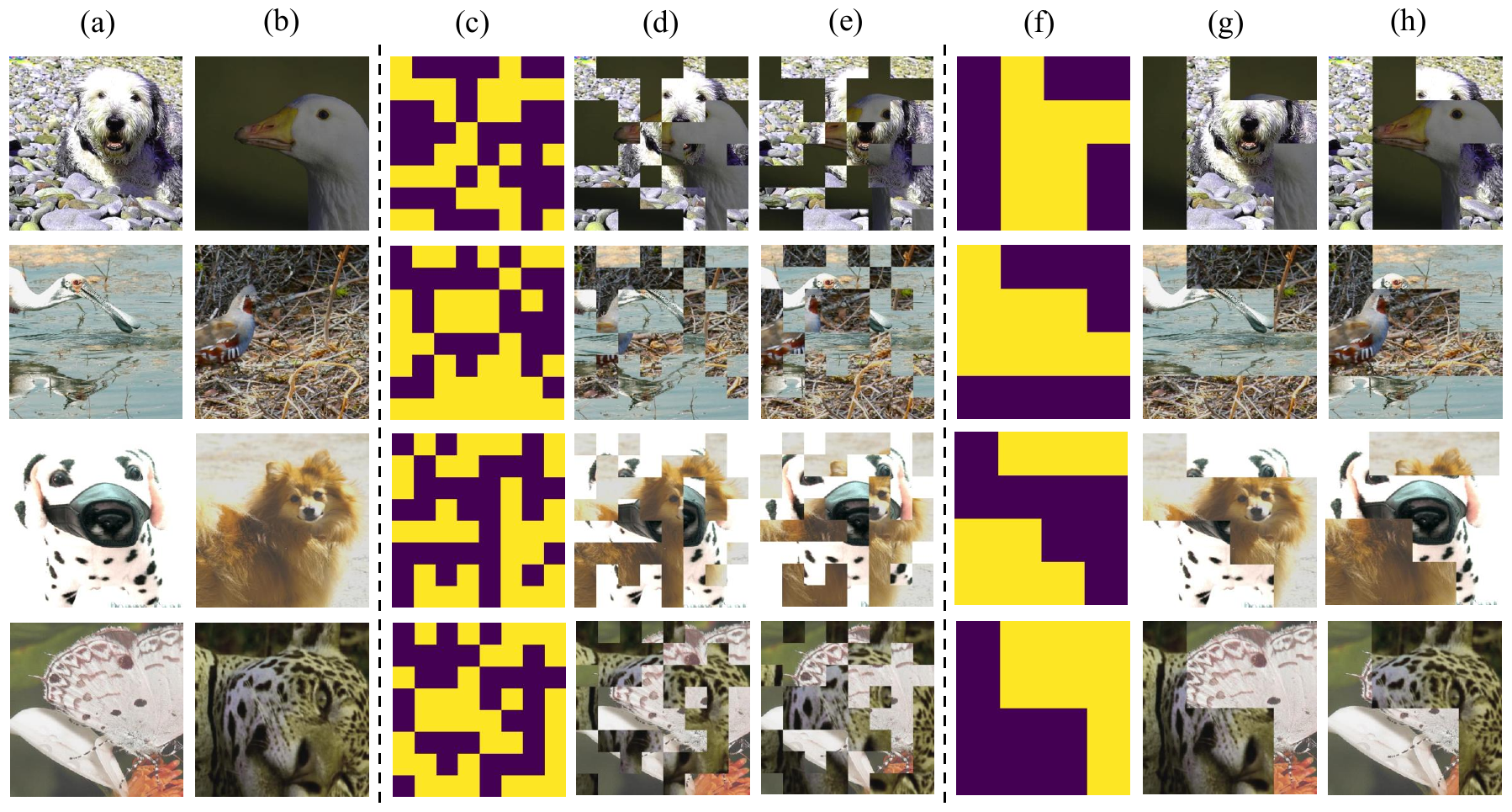}
  \caption{Illustration of the different mask patterns with a mask grid size of 8. (a) and (b) are input images. (c) is the discrete/random mask pattern, and (d) and (e) are mixed images using this mask. (f) is the blocked mask pattern, and (g) and (h) are mixed images with a blocked mask. {Discrete masking breaks (c) -- (e) the completeness of an object which is important for the contrastive loss because it operates on the global object level. On the other hand, blocked masking (f) -- (h) preserves important global features leading to superior performance.}}
  \label{fig:masking_appendix}
\end{figure*}

We provide additional illustrations for the different mask patterns and images generated by them. In Fig. \ref{fig:masking_appendix} illustrations we use mask with grid size 8. All original images are sampled from ImageNet-1K.

\end{document}